%% file: KDD23.tex
\documentclass[sigconf]{acmart}
\usepackage{CJKutf8}
\usepackage[utf8]{inputenc}
\input{zhwinfonts}
\pdfoutput=1
\renewcommand\footnotetextcopyrightpermission[1]{} 
\usepackage{fancyhdr} %ok
\usepackage{microtype}      % microtypography
\usepackage{url}  %Required
\usepackage{graphicx}  %Required
\usepackage{color}
\usepackage{bm}
\usepackage{stfloats}

\usepackage{enumitem}
\usepackage{multirow}
\usepackage{soul}
\usepackage{algorithm}  
\usepackage{algpseudocode} 
\usepackage{amsmath}
\usepackage{makecell}
\usepackage{stfloats}
\usepackage{float}
\usepackage{graphicx}
\newcommand{\figref}[1]{Fig.~\ref{#1}}

\newcommand{\secref}[1]{Sec.~\ref{#1}}

\newcommand{\modelname}{\textit{MEBART}}
\AtBeginDocument{%
  \providecommand\BibTeX{{%
    \normalfont B\kern-0.5em{\scshape i\kern-0.25em b}\kern-0.8em\TeX}}}

\setcopyright{acmcopyright}
\copyrightyear{2023}
\acmYear{2023}
\acmDOI{XXXXXXX.XXXXXXX}

\acmConference[xxx]{xxx}{xxx}{xxx}
\acmPrice{15.00}
\acmISBN{978-1-4503-XXXX-X/18/06}

\begin{document}

%\title{Personalized and Trend-Aware Headline Generation for Posts on Social Media}
%\title{How to Generate a Popular Post Headline on Social Media? \protect\\ 
%Keep Your Style yet Chase Trends!}
\title{How to Generate Popular Post Headlines on Social Media?}

\author{Zhouxiang Fang$^{\dagger}$, Min Yu$^{\S}$, Zhendong Fu$^{\dagger}$, Boning Zhang$^{\dagger}$, Xuanwen Huang$^{\dagger}$, Xiaoqi Tang$^{\S}$, Yang Yang$^{*\dagger}$}
\affiliation{%
\institution{$^{\dagger}$Zhejiang University, $^{\S}$Huayun informaton and technology co. Ltd}
\city{Hangzhou}
\country{China}
}
\affiliation{%
\institution{
\{zhouxiangfang, zhendongfu, zhangbn, xwhuang,  yangya\}@zju.edu.cn\\ 
graceyu123@hotmail.com\\
932947388@qq.com}
\city{Hangzhou}
\country{China}
}
\thanks{*Corresponding author}

% \author{Zhouxiang Fang}
% \email{zhouxiangfang@zju.edu.cn}
% \author{Min Yu}
% \email{graceyu123@hotmail.com}
% \author{Zhendong Fu}
% \email{zhendongfu@zju.edu.cn}
% \author{Boning Zhang}
% \email{zhangbn@zju.edu.cn}
% \author{Xuanwen Huang}
% \email{xwhuang@zju.edu.cn}
% \author{Xiaoqi Tang}
% \email{932947388@qq.com}
% \author{Yang Yang}
% \email{yangya@zju.edu.cn}

\begin{abstract}
% With the development of the Internet, more and more users are willing to share their lives and content creations on social media. Among these content creations, there is some widely disseminated content.  High-explosive posts have huge social influence and contain huge commercial potential , which have been studied in various fields such as computer, literature, and psychology. The headline is one of the most important cores of high-explosive posts.  However, generating headlines is difficult. Manually writing requires extensive effort of human to promise quality. Therefore, using automatic methods to generate headlines is promising. However, current methods ignore the characters of social media. To better understand the popular headlines on social media, we collect over one million posts on Xiaohongshu, which is a famous content platform in China, and conduct a careful observation. The results demonstrate that personal styles and trends and are key characteristics of the popular headlines. Inspired by these conclusions, we propose a Personalized and Trend-Aware Headline Generation method. The experiment results show that our proposed model achieves SOTA performance on a real-world dataset, and is aware of the personalized features of users and current hot trends on the platform.
Posts, as important containers of user-generated-content pieces on social media, are of tremendous social influence and commercial value. As an integral components of a post, the headline has a decisive contribution to the post's popularity. However, current mainstream method for headline generation is still manually writing, which is unstable and requires extensive human effort. This drives us to explore a novel research question: Can we automate the generation of popular headlines on social media?  We collect more than 1 million posts of 42,447 celebrities from public data of Xiaohongshu, which is a well-known social media platform in China. We then conduct careful observations on the headlines of these posts. Observation results demonstrate that trends and personal styles are widespread in headlines on social medias and have significant contribution to posts' popularity. Motivated by these insights, we present \modelname\footnote{\url{https://anonymous.4open.science/r/MEBART-AEE2/}}, which combines \textbf{M}ultiple preference-\textbf{E}xtractors with \textbf{B}idirectional and \textbf{A}uto-Regressive Transformers (BART), capturing trends and personal styles to generate popular headlines on social medias. We perform extensive experiments on real-world datasets and achieve state-of-the-art performance compared with several advanced baselines. In addition, ablation and case studies demonstrate that \modelname~advances in capturing trends and personal styles. 
\end{abstract}


\keywords{social media, data mining, headline generation}


\maketitle

\input{introduction}
\input{preliminaries}

\input{observation}
\input{methods}
\input{experiment}

\input{related}
\input{conclusion-and-future-work}


\appendix
\input{appendix}

\end{document}

%% file: introduction.tex
\section{introduction}

\begin{figure}[ht!]
	\centering
	\includegraphics[width=0.40\textwidth]{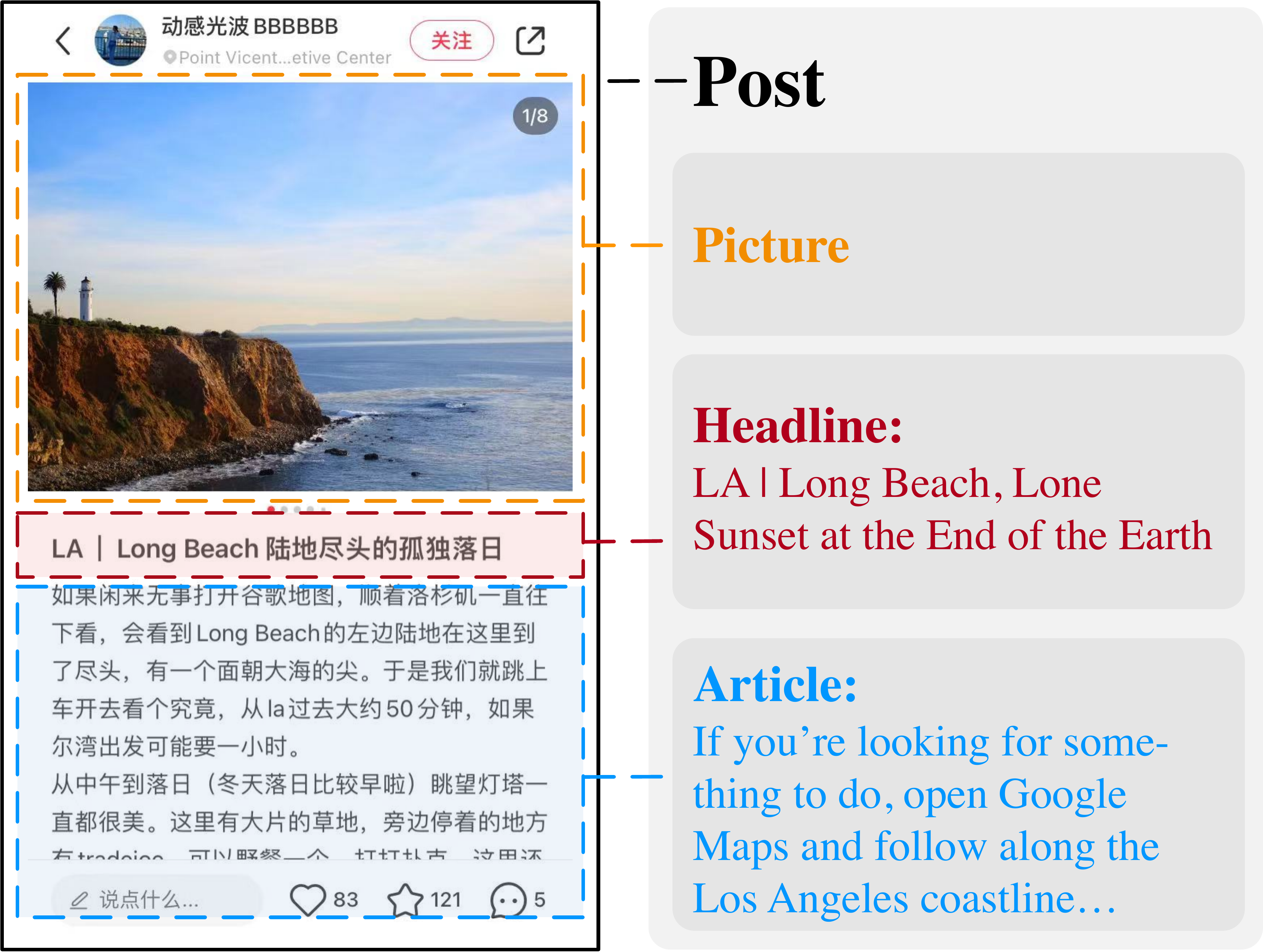}
	\caption{A specific case of a popular post on Xiaohongshu.
	\normalsize
	}
	%and new tasks arriving. When a new task arrives, this method creates a new GNN, which learns the task under the current graph structure..}
	\label{fig:intro}
\end{figure}

% As the Internet has developed, social media become an integral component of our lives \cite{}.
% Every day, billions of pieces of user-generated content, such as opinions, life stories, and experiences, are distributed on social media \cite{saravanakumar2012social}. 
% For instance, in 2022, Twitter had 217 million active users who sent 500 million tweets daily\footnote{\url{https://www.omnicoreagency.com/twitter-statistics/}}.
% On the majority of social media platforms, such as Facebook, Instagram, and Xiaohongshu, those large amounts of user-generated content are distributed via posts.
% Some posts on the social media platform can become popular and be read by a significant number of users.
% These popular posts have a tremendous social impact for a wide range of applications, including government policy propagandizing \cite{}, product marketing \cite{}, etc. Thus, creating popular posts has huge commercial and research values. It has been investigated by many works, involving various academic areas, such as literature \cite{}, marketing \cite{}, and computer science \cite{}.
As the Internet develops over time, social medias have become an integral component of our lives \cite{mcmillan2006coming}.
Every day, billions of user-generated-content pieces, such as opinions and experiences, are presented on social medias \cite{saravanakumar2012social}. 
The post is a basic container of user ideas \cite{morrison2013posting}. 
\figref{fig:intro}~shows a post on Xiaohongshu\footnote{\url{https://www.xiaohongshu.com/}}, a well-known social media platform in China. 
Due to its huge reads, some popular posts help exert a tremendous influence for a wide range of applications, including government policy propagandizing \cite{bradshaw2020propagandizing}, product marketing \cite{de2012popularity}, etc. 
Therefore, creating popular posts is of huge commercial and research value. 
%It has been investigated by many works, involving various academic areas, such as literature \cite{}, marketing \cite{}, and computer science \cite{}.

Creating a popular headline is a fundamental part of creating a popular post. According to the literature \cite{lakkaraju2013s, zagovora2018increases}, a headline, as a crucial part of a post \cite{meij2012adding}, has a decisive influence on the post's popularity. A successful headline can raise the post's popularity (reads) by drawing more readers' attention. Besides, the headline summarizes the central point of a post, the quality of which might impact the readers' initial impressions. 

% However, how to make eye-catching headlines on social media is a challenging problem. Deciding which aspects of the content to present in the title and how to write the title requires the creator have cognition of current user aesthetics and solid writing skills. Meanwhile, varying user interests, and social trends also further add to the complexity of this problem. At present, the main method of headline generator is still manual writing. However, it requires extensive human effort to promise the quality of the generated headline. The writer requires to learn writing skills, understand the content, and be aware of current trends. Meanwhile, the self-style should be also toke into account in social media. \textbf{Therefore, this paper explores a novel research question, can we automate the generation of popular headlines on social media?}

In the past, individual users usually create headlines by themselves. However, it is hard to guarantee the popularity of the generated headlines due to the large variation in individual knowledge and personal properties. 
Recently, some companies (e.g., the Multi-Channel Network\footnote{the Multi-Channel Network (MCN) is any entity or organization which either partners with content creators or directly produces a variety of distinctive content}) notice the massive market of this business opportunity and provide professional writing and ghostwriting service for clients \cite{gardner2016s}. 
However, due to users' interests and social trends, this requires extensive human effort. 
What's more, its high service fee hinders itself from becoming prevalent among users.
Therefore, this paper explores a novel research question:  \textbf{Can we automate the generation of popular headlines on social media?}

% However, writing a popular headline on social media is complex, involving many factors.
% Firstly, extracting and summarizing the fascinating key points from a post's content are fundamental requirements for a popular headline; hence, a thorough understanding of the post's content and proficient writing abilities are essential. 
% Second, as it is common for social media users to follow trends, consistency with trends can lead to increased media exposure for headlines. Consequently, the design of headlines also necessitates an awareness of current trends and the proper integration of trend components.
% Thirdly, certain social media users have unique and influential styles that could attract a great deal of attention, and headlines are an excellent way to highlight them. Therefore, it is essential on social media to generate headlines that consider personal style or celebrity style.
Currently, with the advancement of the sequence-to-sequence technique in Natural Language Processing (NLP) \cite{vaswani2017attention, devlin2018bert, lewis2019bart}, headline generation methods have been studied extensively \cite{liu2022brio, lewis2019bart} and achieved a great success in a variety of scenarios, such as news writing \cite{ao2021pens}, online ads \cite{kanungo2022cobart}, and paper writing \cite{shenassa2022elmnet}. 
Most of methods focus on a single document (an article and a headline), and aim to extract and summarize the text's key points.
In other words, their input is a specific text and output is the corresponding headline.
However, generating headlines on social media is a more complicated issue. 
It is common for social media users to follow trends \cite{wang2019hot}, since keeping up with trends can enhance headlines' exposure on the media platform. Hence, headline generation should intuitively consider current trends.
Besides, some of the users have their unique and influential styles that attract a great deal of attention \cite{liu2017self}, which is highlighted in effect headline creating.
Therefore, personal style is also essential for headline generation on social media. 
Considering these key factors, directly extending current methods of headline generation on social media may not achieve a satisfactory performance. 

However,  it is challenging to include trends and personal styles in headline generation.
First, it takes large-scale datasets to study this issue. 
% \fzx{challenge1}
Furthermore, it is difficult to figure out how trends and personal properties affect the generation process and popularity of headlines. 
To the best of our knowledge, there is no study exploring these issues on large-scale and real-world datasets. 
Lastly, since trends and personal styles always change over time, how to encode and combine them is also a great problem. 
% \fzx{challenge3}

To overcome these challenges, this paper collected more than 1 million posts from 42,447 thousand from  public data of Xiaohongshu. We conduct careful observations on the headlines of these posts. 
Results indicate that trends and personal styles indeed exist in the headlines of posts and contribute significantly to posts' popularity. 
Inspired by these observations, we present \modelname, which combines \textbf{M}ultiple preference-\textbf{E}xtractors with \textbf{B}idirectional and \textbf{A}uto-\textbf{R}egressive
\textbf{T}ransformers (BART) \cite{lewis2019bart} to model trends and personal styles for headline generation on social media. 
Specifically, the preference-Extractors respectively encode the users' previous headlines and buzzwords generated from previous headlines, then output the corresponding preference encodings. 
Then these encodings are combined with the hidden states of the articles to generate the final headlines. 
To improve the ability of preference-Extractors of \modelname, we design a novel denoising-based pretraining strategy for them. 
In the pretraining stage, the target headline is first corrupted with token masking and replacing. \modelname~ then combines the corrupted headline,  corresponding style encoding and trend encoding to reconstruct the original target headline.

Extensive experiments of headline generation on a real-world dataset of Xiaohongshu demonstrate the effectiveness of our proposed model: \modelname~ significantly outperforms several state-of-the-art baselines (average \textbf{9.9\%} in term of ROUGE-L). 
We also conduct in-depth ablation studies and quantify different components' contribution to the performance of \modelname, showing the effectiveness of combing trends and personal style. 
Several translated cases are displayed to illustrate that \modelname~advances in capturing trends and personal styles.
% \fzx{experiment:strong model;ablation:intuition;cases:do capture trends and style;}

Overall, the contributions of this paper can be described as follows:
\begin{itemize}[leftmargin=*]
    \item We present a novel study: automatically generate popular headlines on social media; and investigate the problem from the perspectives of trends and personal styles.
    \item Comprehensive observations of over 1 million posts from Xiaohongshu illustrate the importance of trends and personal styles for creating popular headlines.
    \item Motivated by our observational insights, we propose \modelname, with a novel pretraining strategy, for headline generation. Experiment results on real-world datasets validate the effectiveness of our proposed method.
%Furthermore, detailed case studies illustrate various diffusion patterns of specific products, which provide additional insights that will help to facilitate a full understanding of user online purchasing behaviors.
\end{itemize}

% A key part of the operation of the Internet celebrity economy is the creation of high-exposure (popularity) content. On the one hand, it can help Internet celebrities enhance their own influence, and on the other hand, they can better promote commercial content. From common sense, as well as relevant literature, a good title is a key factor in determining the content of high exposure.

% However, creating a popular title is a challenge task. 
% ``popular'' depends on human senses, there is not a precise definition that a popular title should contains what. 
% Therefore, at present, the most mainstream method of title retrieval is still manual retrieval. 
% E.g., Professionals from Internet celebrity incubators. 
% This need a huge human resources and professional background. 
% \textbf{Therefore, this paper investigate a new possible, can we automatically generate popular titles via deep neural networks?}

%Title generation is a classic sequence-to-sequence problem in NLP. In the past, many works are devoted in title generation. Some of them target for generating summerize-text for a content, such as: A, B, C\cite{}. In news media, there are aslo some works foucus on popularity title generation. However, how to generate popular title in social media are still not be investigate. 

%The most feature of popular title social media than other title is, in social media, the posts are published by a person, with a specific publish time. In this paper, we carefully observe more than 1M posts on Xiaohongshu, which is a popular content community in China. 

%% file: preliminaries.tex
\section{preliminaries}
For post $z_i$, $x_i^{ar}$, $x_i^{se}$, $x_i^{te}$ and $y_i$ are respectively the article, style text, trend text and headline. Time step $t_x$ is the $x$th month-grained time step in increasing order. In other words, $t_x = t_{x-1}+1$. $T_{z_i}$ is the time step of $z_i$, demonstrating the month when the post is released. $Y^{t_x}$ is all the headlines posted within time step $t_x$. $tf(w,t_x,V)$ is the token frequency of token $w$ in vocabulary $V$ for all the headlines within time step $t_x$. $tf(t_x,V)$ refers to all the token frequencies in $V$ within time step $t_x$. $BW(t_x,V)$ is the buzzwords list of $t_x$ in vocabulary $V$. $W_{z_i}$ denotes the author(user) of $z_i$ and $Z_{u_a}$ denotes the posts list of user $u_a$.

%% file: observation.tex
\section{observation}

\begin{figure*}[!t]
\centering
\includegraphics[width=1.0\linewidth]{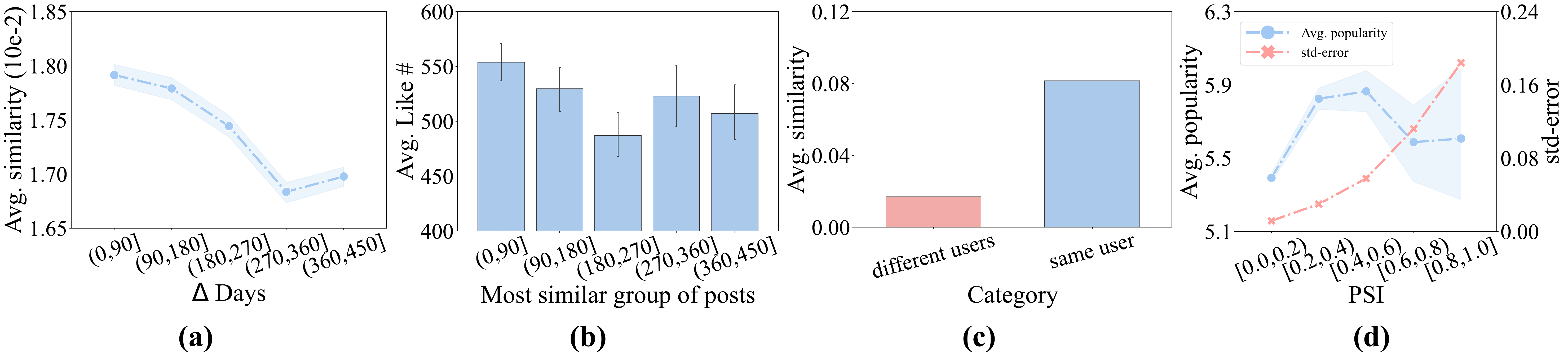}
\caption{Observations of the trends and personal styles of headlines. (a) shows the average similarities between the headlines from different seasons. (b) shows the average likes of headlines from different season groups. A headline is distributed to the season group the similarity between which is the highest among all.(c) shows the average similarities between the headlines of different users and the same users. (d) shows the average popularity of users with different extent of personal styles.}
\label{fig:analysis}
\end{figure*}

In this section, we conduct in-depth analyses of headlines on social media platform. We mainly focus on two key factors for creating headlines: personal styles and trends. First, we introduce the dataset used for the analyses. Then, for each factor, we explore whether it is widespread in the headlines of posts and measure its contribution to popularity of posts.

\subsection{Dataset Descriptions}\label{dataset}
The dataset for our experiments is described as follows: we sample posts on Xiaohongshu during a specific period of time from 2020 to 2022. The raw dataset contains 1,695,219 posts and 42,447 users. Each post consists of basic information, including: 1) the article; 2) the headline; 3) received likes; 4) when the post is posted, namely time step; and 5) who creates the post, namely the user. 
% (1) 使用了2021年全年的数据 
% (2) 为了避免太多无意义的post的影响，我们只选取具有一定流行度的post(点赞量大于10)
% (3) 为了统计数据不受用户发布post的数量影响，我们选取每个用户发布时间最新的20篇post 

\subsection{Analyses of Trends}
\label{obs:Q1}
% ----------Question one----------
% 有热门话题->热门话题在变化->社交媒体上的用户follow这种变化->标题随着时间变化
% 在Google trends上，他展示了互联网上每一天都会出现大量新的热搜词语。通常来说，这些热搜词语反映了一段时间内的各界大事与流行话题,它的特点是随着时间变化，并且可以吸引大量的受众关注。通过观察小红书的数据，我们发现社交媒体平台上的用户也会跟随这些热门话题，而改变自己的文章的主题。而标题作为一个文章内容的总结提炼，最能反映出这篇文章所讨论的主题。为了仔细的研究这个现象，我们定义了问题1
% ----- ----- ----- ----- ----- ----- ----- -----

\textbf{Existence of Trends}

A trend is a change or development towards something new or different, which can be revealed in many aspects, including hot topic, buzzwords and prevalent sentence patterns. 
Take buzzwords for example, the Google Trends\footnote{\url{https://trends.google.com/trends/}} has shown that lots of buzzwords appear on the Internet every day. 
However, it remains opaque that whether trends is widespread in social medias as well. Specifically, does a user take trends into consideration when he or she creates a headline? Therefore, we define the first research question: \textbf{Are users' headlines influenced by trends on social media? (Q1)}

To address this question, We first sampled 50,000 headlines from the last 10 days of 2021 as our study object, namely $Y^{ob}$. 
The reason we choose a certain time period for analysis is to control for extraneous variables, such as posting frequency, festival and platform support. 
We have also conducted analysis on other time periods in Appendix \ref{time}.
% \fzx{other time period}
Then we sample 500 headlines for each of five different seasons. 
Each season is 90 days long and adjacent to its precedent. 
The first season is closest to $Y^{ob}$.
Finally, we calculate the average text similarities between headlines from $Y^{ob}$ and different seasons.
In other words, we measure the similarity between headlines from the last 10 days of 2021 and from different seasons before the last 10 days of 2021.

\citet{yan2021chinese} has demonstrated that buzzwords are crucial for online interpersonal communication. 
Therefore, we use \textit{Jaccard similarity} as the metric of text similarity, since it's simple and capable to detect buzzwords. 
% \fzx{why jaccard}
The Jaccard similarity can be clarified as:
\begin{equation}
    \mathbf{SIM}(y_i, y_j) = \frac{|set(y_i) \cap set(y_j)|}{|set(y_i) \cup set(y_j)|},
\end{equation}
where $y_i$ and $y_j$ denote two different headlines and $set(\cdot)$ is the set operation that removes duplicate words. 
Besides, we have also conducted analysis with other metrics of text similarity \cite{gomaa2013survey} in Appendix \ref{metric}. 
% \fzx{other metirc}
The higher the $\mathbf{SIM}$, the more similar the two headlines are.

The results are shown in \figref{fig:analysis}~(a): the average $\mathbf{SIM}$ reaches the maximum of $0.0179$ in the most recent season and the minimum of $0.0168$ in the fourth most recent season. 
In general, the $\mathbf{SIM}$ decreases as the difference of publishing time increases. 
However, when the difference of publishing time reaches a year, the $SIM$ starts to rise back instead. The reason lies behind this may be the cyclical nature of trends~\cite{robinson1975style}. 
We notice that the frequencies of some words, such as Christmas and Chinese New Year, are strongly  associated with the time. 
Besides, although some words are not directly related to time, such as a festival or anniversary, we can still observe significant rise of frequencies of them. 
We think these words can reveal how would the users prefer to describe things, contemporary prevalent topics and other properties that are related to trends. 

To sum up, based on the analysis above, we conclude that headlines are influenced by trends. In other words, \textbf{trends do exist in headlines on social medias}.

% -----Question two-----
% 承上启下 -> 受众喜欢新的内容 -> 我们猜测越新的内容，文章流行度越高
% 通过上面的分析，它展示了在小红书数据上同样存在trends，并且用户通过改变自己的文章去追随当下的热门话题。通常来讲，受众会更喜欢选择这一类和当下热门话题相关的文章去浏览和点赞。因此，我们猜测可能越潮流的文章就有具有更高的流行度。
% ----- ----- ----- ----- ----- ----- ----- -----
\textbf{Trends Influence Popularity}

The above analysis demonstrates the existence of trends on Xiaohongshu, which influence the creating of headlines for many users . 
However, it remains unknown that whether following trends help raise popularity of the posts. 
Therefore, we introduce the second research question: \textbf{What effect does the trends exert on the popularity of the posts? (Q2)}

% 为了研究这个问题，我们对比在同一个时间段内发布的文章不同潮流对流行度的影响。我们延续上面的做法，把2021年最后10天的文章作为我们的研究对象。为了简化问题，我们粗糙的认为不同的季度拥有不同的潮流。所以我们想找到 研究对象中的每一篇文章最符合哪个季度的潮流。具体来说，我们分别计算一篇文章和哪个季度的文章最相似，那么这篇文章就符合这个季度的潮流。
To investigate this question, we continue to use the same $Y^{ob}$, separated seasons and metric of similarity above. 
Each headline in $Y_{ob}$ are marked by the season in which the headlines are most similar to it.
Then the headlines are divided into five groups according to their marks.
Finally, We calculated the average number of likes for each group. 

The results are shown in \figref{fig:analysis}~(b): the average likes reaches the maximum of $553.78$ in the most recent season and the minimum of $486.93$ in the third most recent season. 
In general, as the difference of publishing time increase, the number of likes decreases. 
But when the difference of publishing time reaches one year, the number of likes rises again.
Since the difference of publishing time is roughly negative correlated with the extent of following trends as shown in the previous chapter, these results have shown the positive correlation of the extent of following trends and received likes.

To conclude, \textbf{a headline which follows trends has positive influence on the popularity of the post.}

\subsection{Analyses of Personal Styles}
% ------e----Question three----------
% characteristic of data/application -> phenomenon -> intuition -> observation (maybe driven by problem, maybe directly observe)
% ----- ----- ----- ----- ----- ----- ----- -----
\textbf{Existence of Personal Styles} 

In the field of literature, many writers have their own styles of writing. 
For example, the style of Hemingway is recognized as abbreviated and simple while that of Jane Austen is recognized as fine and detailed. 
However, do personal styles exist in social medias as well?
Social media platforms such as Xiaohongshu have abundant users and each of them may has his/her own style.
Therefore, we introduce the third research question: \textbf{Do personal styles exist in creating headlines on social medias? (Q3)} 

% 根据前面所述，不同的用户具有不同的语言风格，这可能导致不同用户在写文章的时候，有不同的用词习惯。因此，为了仔细的研究每个作者的标题是否具有个人风格，我们对比来自于相同作者的标题对和来自于不同作者的标题对之间文本相似度分布的区别。
To address this question, we compare text similarities between the headlines from the same users and different users.
% 我们采样100万个标题对，这里的每个标题对里面的两个标题来自于同一作者，并计算这来自于同一作者的标题之间的文本相似度的期望。
Specifically, we first sample $1,000,000$ headline pairs, in which the two headlines are from the same user, and calculate the expectation of text similarity between headlines for all the pairs.
% 使用相同的做法计算这来自于不同作者的标题之间的文本相似度的期望，除了每个标题对里面的两个标题来自于不同的作者。
Then we do the same for the headline pairs in which the two headlines are from different users.
As in \secref{obs:Q1}, we use \textit{words overlap} as the metric of text similarity. 

% 展示图表，进行分析
The results are shown in \figref{fig:analysis}~(c): 
the average $\mathbf{SIM}$ of the headline pairs from the same users is $0.081$, much higher than that ($0.017$) of the headline pairs from different users. 
This demonstrates that headlines of the same users are more similar compared to the headlines of different users. 
To better understand this result, we observe several cases and find out a user is inclined to imitate his or her old headlines when making a new one in many aspects, including the typesetting patterns and choices of words.

In other words, \textbf{personal styles do exist in headlines on social medias.}

% ----------Question four----------
% ----- ----- ----- ----- ----- ----- ----- -----

\textbf{Personal Styles Influence Popularity} 

Although we are informed of the existence of personal styles, its relationship with post popularity remains unknown. 
Just like everyone has a different taste for books, readers may be attracted by headlines of different styles. 
Intuitively, users with apparent personal style may be more likely to be identified by the readers, thus drawing more attention. 
To investigate the relationship between personal styles and post popularity, we introduce the fourth research question: \textbf{What effect does the personal styles in headlines have on post popularity? (Q4)}

First, to quantify the personal styles, we introduce the \textit{Personal Style Index} ($PSI \in [0,1]$), which denotes the extent of personal style in the headlines of a user. Calculating $PSI$ for user $u_a$ is defined as:
\begin{equation}
    \mathbf{PSI}(u_a) = \frac{\sum_{y_i \in Y_{u_a}} \sum_{y_j \in Y_{u_a}} \mathbf{SIM}(y_i, y_j)}{|Y_{u_a}|^2},
\end{equation}
where $Y_{u_a}$ denotes the posted headlines of user $u$. Higher $PSI$ demonstrates that the user has a more distinctive personal style. When $PSI$ is equal to $1$, it means that all the headlines of the user are exactly the same.

Then, to quantify the popularity of users, we introduce the \textit{popularity index} ($PI$) as follows:
\begin{equation}
    \mathbf{PI}(u_a) = \frac{\sum_{z_i \in Z_{u_a}} log(likes(z_i))}{|Z_{u_a}|},
\end{equation}
where $Z_{u_a}$ denotes all the posts of user $u_a$ and $likes(z_i)$ denotes the number of likes of post $z_i$.

% 展示图表，进行分析
Finally, we divide the users into five groups according their $PSI$, and calculate $PI$ for each group. The results are illustrated in \figref{fig:analysis}~(d): the $PI$ of all users is $5.43$, reaching the maximum of $5.86$ when $PSI$ is in the interval of $[0.4, 0.6)$ and the minimum of $5.39$ when $PSI$ is in the interval of $[0.0, 0.2)$. 
In general, as the $PSI$ increases from $0$ to $1$, the average popularity of users first increases and then decreases. 
These results suggest that posts with stronger personal styles are generally more popular compared to those without personal styles. 
However, if the personal style are too strong, the popularity may drop instead.
The reason lies behind the drop may be that as the proportion of fixed words in the headlines becomes too large, there are few useful information for the readers in the headlines, which discourage them from clicking into the post. 

We also calculate the standard error of popularity for each interval shown in \figref{fig:analysis}~(d): the standard error increases from $0.011$ to $0.184$ as $PSI$ becomes larger. 
This demonstrates that popularity becomes quite extreme when personal style is too strong. 
For big influencers, using almost the same headlines may help them constantly make connections with their old fans. 
On the other hand, similar headlines seems kind of boring and may push the new readers away, which is not a good choice for small influencers who are eager to increase their fans. 
% \fzx{explain standard error}

To conclude, \textbf{personal styles within appropriate extent exert positive effect on the popularity of posts.}

%% file: methods.tex
\section{method}

\begin{figure*}[!t]
\centering
\includegraphics[width=1.0\linewidth]{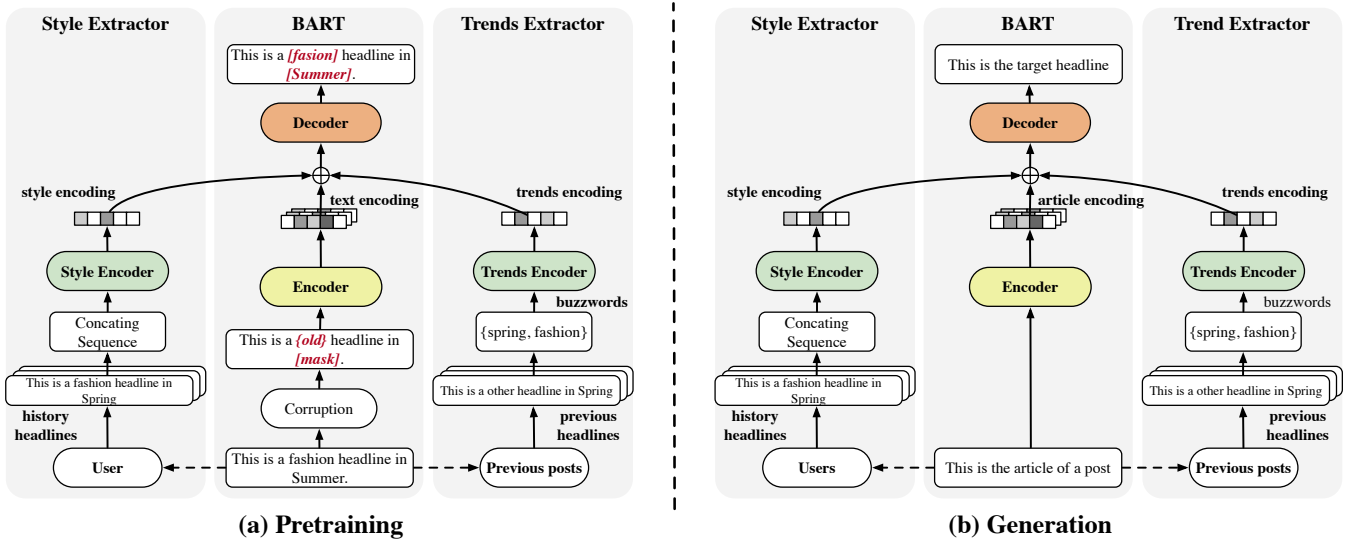}
\caption{The overview of \modelname. (a) displays the pipeline of \modelname~during the pretraining stage, and (b) illustrates the pipeline of \modelname~during the generating(including training and testing) stage.}
\label{fig:model}
\end{figure*}

\subsection{Overview of MEBART}
The observations have demonstrated the existence of personal styles and trends, as well as their influences on popularity of posts on social medias. 
Inspired by these, we propose \modelname~ for generating personalized and trend-aware headlines. 

In general, \modelname~ is a BART-based headline generator composed of four components, namely Encoder $f_{en}(\cdot)$, Decoder $f_{de}(\cdot)$, Style Extractor $f_{se}(\cdot)$ and Trend Extractor $f_{te}(\cdot)$. 
The architecture of \modelname~ is shown in \figref{fig:model}.
Like most of Transformer-based auto-regressors, firstly $f_{en}(\cdot)$ encodes the article into a sequence of hidden states $H_i^{ar} = (h_{i1}^{ar},h_{i2}^{ar},...,h_{i{n_i}}^{ar})$, then $f_{de}(\cdot)$ takes $H_i^{ar}$ as input to generate a headline. 
Based on this framework, we introduce two kinds of preference-Extractors ($f_{se}(\cdot)$ and $f_{te}(\cdot)$) to model styles and trends respectively. 
The preference-Extractor takes representative text of the author's style or contemporary trends as input, and generate corresponding encoding as output. 
For an article $x_i^{ar}$, $h_i^{se}$ and $h_i^{te}$ respectively denote the \textbf{style encoding} and \textbf{trend encoding}. 
Since the qualities of $h_i^{se}$ and $h_i^{te}$ are crucial to the final generated headline, we propose a self-supervised learning task to pretrain $f_{se}(\cdot)$ and $f_{te}(\cdot)$. 
During generation, $h_i^{se}$ and $h_i^{te}$ are combined together to modify $H_i^{ar}$, guiding the model to generate personalized and trend-aware headlines. 
Details of how each component functions and the whole working pipeline of \modelname~ are described below. 

\subsection{Modeling Personalized-styles and Trends}

\textbf{Collecting Posted Titles and Buzzwords}\label{chap:collect}

The intention of the preference-Extractors is to model the ``preferences'' in styles and trends.
As mentioned in the overview, for an article $x_i^{ar}$, the preference-Extractors require representative text  (\textbf{style text} $x_i^{se}$ and \textbf{trend text} $x_i^{te}$) as input. 
The qualities of $x_i^{se}$ and $x_i^{te}$ largely determine the qualities of $h_i^{se}$ and $h_i^{te}$, and indirectly influence the model's ability to capture styles and trends. 
However, it is challenging to form $x_i^{se}$ and $x_i^{te}$ that can truly \textbf{speak for} styles and trends, since they are quite complex and integrated concepts.
In other words, we can hardly define styles and trends from plain intuition. 
Therefore, we carefully design the strategies for constructing $x_i^{se}$ and $x_i^{te}$. 

For personal styles, we have be informed by the observation that many users have their own styles in creating headlines, which help them quickly draw the attention of their ``old fans''.
Therefore, we use the posted headlines of the same user to form $x_i^{se}$, as a reference for the new headline to be generated. 
For post $z_i$ and its author $u_a$, $x_i^{se}$ is formatted by concating the headlines whose corresponding post $z_j$ satisfies $z_j \in Z_{u_a}$ and $T_{z_j} < T_{z_i}$ in decreasing order of time step. 

For trends, the observation has shown that the similarity between headlines increases as the difference of their publishing time decreases, which supports the existence of trends. 
The observation also demonstrates that following trends helps exert positive effect upon the popularity of posts. 
However, different from personal styles, directly concating all the headlines within $t_x$ to form trend text for all the article within $t_x$ is unreasonable. 
Considering the amount of $Y^{t_x}$, this approach may introduce abundance of duplicated and noisy text.
Within limited training time, it's too hard for the model to learn. 
To effectively represent the trends, we use buzzwords\cite{nakajima2012early} instead, which are prevalent words during a certain period. 
The first way of selecting buzzwords that comes to mind may be sorting words by their token frequency (tf). 
However, it is not appropriate to simply define the buzzwords as words with \textbf{absolutely-high} tf in the moment. 
This kind of definition may ignore some important rising trends, which will be popular in the near future. 
For example, a new APPLE product, a new record of Rihanna, or news that has just been reported, etc.
Therefore, we make a trade-off. 
That is, when collecting the buzzwords, we also pay attention to those words with \textbf{relatively-high} tf compared to the past.

To be specific, we first collect headlines in the train dataset and use them to generate a headline vocabulary $V$. 
Words are filtered with minimum tf ($tf_{min}$) and maximum tf ($tf_{max}$) . 
Then we calculate $tf(t_x,V)$ for every time step $t_x$. 
In real-world, users can only take the headlines in the past as references for creating a new headline.
Therefore, to generate buzzwords list for $t_x$, we can only use the information of tf before $t_x$. 
In other words, $BW(t_x)$ should be produced with $tf(t_y,V)$ where $t_y < t_x$ only. 
So $BW(t_0,V)$ is actually empty. 
Finally, tokens with different time-grained relatively-high tf and absolutely-high tf are selected in balance. 
All the articles within $t_x$ share the same $BW(t_x,V)$, which will be concated to form the same trend text during generation. 
Detailed algorithm of generating $BW(t_x,V)$ can be formulated as in Algorithm \ref{alg}. \\ 

\begin{algorithm}[b]  
  \caption{Generate Buzzwords}
  \label{alg}  
  \begin{algorithmic}  
    \Require  
        Time step: $t_x$; Headlines list: $\{Y^{t_0}, Y^{t_1}, Y^{t_2}...\}$;Headline vocabulary: $V$;
    \Ensure 
 		Buzzwords list $BW(t_x,V)$ for time step $t_x$
 	\If {$t_x \geq 2$} 
        \For{$w \in V$}
            \State $R(w) \leftarrow  tf(w, t_{x-1},V) / tf(w, t_{x-2}, V)$;
        \EndFor
        \State Add the 128 tokens with highest $R(w)$ to $BW(t_x,V)$
    \EndIf
    
    \If {$t_x \geq 6$}
 		\For{$w \in V$}
            \State $R(w) \leftarrow  tf(w,(t_{x-1~to~x-3}),V) / tf(w,(t_{x-4~to~x-6}),V)$;
        \EndFor
        \State Add the 64 tokens with highest $R(w)$ to $BW(t_x,V)$
    \EndIf
    
    \If {$t_x \geq 12$} 
 		\For{$w \in V$}
            \State $R(w) \leftarrow  tf(w,(t_{x-1~to~x-6}),V) / tf(w,(t_{x-7~to~x-12}),V)$;
        \EndFor
        \State Add the 32 tokens with highest $R(w)$ to $BW(t_x,V)$
    \EndIf
    
    \If {$t_x \geq 1$} 
        \State $V' \leftarrow V$ sorted in descending order of $tf(w, t_{x-1}, V)$ 
        \For{$w \in V'$}
            \State Add $w$ to $BW(t_x,V)$
            \If {$sizeof(BW(t_x,V)) = 512$}
                \State break
            \EndIf
        \EndFor
    \EndIf
  \end{algorithmic}  
\end{algorithm}

\noindent\textbf{Preference-Extractor} 

Inspired by \cite{riley2020textsettr}, we propose the preference-Extractor, which is a modification based on the encoder of BART. 
To be specific, the preference-Extractor is composed of two part - a transformer-based encoder\cite{vaswani2017attention} and a pooling layer. The former part is the same as BART's encoder;the latter part is a mean-pooling layer, attached behind the former part.
The encoding sequence of the encoder is turned into a single length-fixed encoding(vector) after pooling.
The dimensionality of the encoding is set to $d$, matched with that of the encoder's hidden states.

Parameters are not shared between preference-Extractors, except the token-embedding layer. 
For $x_i^{ar}$ , $f_{se}(\cdot)$ takes style text $x_i^{se}$ as input and generates style encoding $h_i^{se}$ as output, while $f_{te}(\cdot)$ takes trend text $x_i^{te}$ as input and generates trend encoding $h_i^{te}$ as output. These can be clarified as: 

\begin{equation}
\begin{aligned}
\label{eq:h_i^{se}}
    h_i^{se} = f_{se}(x_i^{se}) 
\end{aligned}
\end{equation}
and 
\begin{equation}
\begin{aligned}
\label{eq:h_i^{te}}
    h_i^{te} = f_{te}(x_i^{te}) 
\end{aligned}
\end{equation}
where the $h_i^{se}$ and $h_i^{te}$ are the results after mean-pooling the last hidden states of the encoder part of the $f_{se}(\cdot)$ and $f_{te}(\cdot)$ respectively. The dimension of them is $d$. \\

\noindent\textbf{Pretraining strategy}

As mentioned before, the intention of preference-Extractors is to capture the \textbf{preference} in styles and trends, eventually helping the model to generate personalized and trend-aware headlines.
Therefore, we design a Masked LM task \cite{devlin2018bert} to improve preference-Extractors' ability.
For $y_i$ and corruption function $\phi(\cdot)$, the corrupted headline is $ \widetilde{y_i} = \phi (y_i)$.
We use the same setting of $\phi(\cdot)$ as \cite{devlin2018bert} does, which includes masking and replacing tokens.
$\widetilde{y_i}$ is then used to predict the original $y_i$. 
To be specific, $\widetilde{y_i}$ is first fed into $f_{en}(\cdot)$ to produce a sequence of hidden state $H_i^{ti} = (h_{i1}^{ti},h_{i2}^{ti},...,h_{i{n_i}}^{ti})$.
Then each hidden state in $H_i^{ti}$ are added up with $h_i^{se}$ and $h_i^{te}$, producing the new hidden state sequence $S_i^{ti}$. 
Finally, $f_{de}(\cdot)$ uses $S_i^{ti}$ to generate the prediction $\widehat{y_i}$. 
These can be formally expressed as:
\begin{equation}
\label{eq:pretrain} 
\begin{aligned}
    H_i^{ti} &= f_{en}(\widetilde{y_i}) \\
    H_i^{se} &= (h_i^{se},h_{i}^{se},...,h_i^{se}) \\
    H_i^{te} &= (h_i^{te},h_{i}^{te},...,h_i^{te}) \\ 
    S_i^{ti} &= H_i^{ti} + H_i^{se} + H_i^{te} \\
    \widehat{y_i} &= f_{de} (S_i^{ti}) \\
\end{aligned}
\end{equation}
where $H_i^{ti}$, $H_i^{se}$ and $H_i^{te}$ has the same length $n_i$. $S_i^{ti}$ is the modified hidden state sequence.  
 
The target of pretraining is to reconstruct the original headline under the ``direction'' of personal styles and buzzwords. The model only predicts masked tokens during pretraining, since other tokens are not corrupted. Therefore, the loss function of pretraining stage can be formulated as follows:
\begin{equation}
\label{eq:pretrain loss} 
\begin{aligned}
    L_{pre} &= -\frac{1}{M_i}\sum_{m=1}^{M_i}\sum_{c=1}^C y_i^{mc} log(\widehat{y_i}^{mc}) \\
\end{aligned}
\end{equation}
where $M_i$ is the number of masked tokens, $C$ is the number of classes (possible tokens). $\widehat{y_i}^{mc}$ is the predicted probability of class $c$ for the $m$th masked token. $y_i^{mc}$ is the real label of class $c$ for the $m$th masked token.  

\figref{fig:model}~(a) illustrates the process of pretraining. During pretraining, $f_{de}(\cdot)$ and $f_{en}(\cdot)$ will be frozen, including the token-embedding layer shared by all the four components of \modelname. 

\subsection{Headlines Generation}

\textbf{Training procedure} 

During training, \modelname~ combines $x_i^{ar}$, $x_i^{se}$ and $x_i^{te}$ together to predict tokens $y_i = (y_i^1, y_i^2,...,y_i^{|y_i|})$.
$x_i^{ar}$ is fed into $f_{en}(\cdot)$ to produce a sequence of hidden state $H_i^{ar}=(h_{i1}^{ar},h_{i2}^{ar},...,h_{i{n_i}}^{ar})$. 
Then each hidden state in $H_i^{ar}$ are added up with $h_i^{se}$ and $h_i^{te}$, which are the encodings of $x_i^{se}$ and $x_i^{te}$, producing the new hidden state sequence $S_i^{ar}$. These can be formulated as follows:
\begin{equation}
\label{eq:train}
\begin{aligned}
    H_i^{ar} &= f_{en}(x_i^{ar}) \\
    H_i^{se} &= (h_i^{se},h_i^{se},...,h_i^{se}) \\
    H_i^{te} &= (h_i^{te},h_i^{te},...,h_i^{te}) \\ 
    S_i^{ar} &= H_i^{ar} + H_i^{se} + H_i^{te} \\
\end{aligned}
\end{equation}
where $H_i^{ar}$, $H_i^{se}$ and $H_i^{te}$ has the same length $n_i$. $S_i^{ar}$ is the modified hidden state sequence.  

The target in the training stage is to predict the original headline with the help of personal styles and trends. Therefore, we define the loss function as:

\begin{equation}
\label{eq:mle loss} 
\begin{aligned}
    L_{tra} &= -\frac{1}{|y_i|}\sum_{q=1}^{|y_i|}logp_{\theta}(y_i^q|y_i^{<q},S_i^{ar}) \\
\end{aligned}
\end{equation}
where $S_i^{ar}$ is the modified hidden state sequence. $y_i^q$ is the $q$th token of $y_i$.

\figref{fig:model}~(b) illustrates the process of training. All the parameters of \modelname~ are trainable in training. \\

\noindent\textbf{Generation deployment}

During generation, we use Beam Search with a beam size of 4 to generate headlines. Beam Search is a searching technique that employs a length-normalization procedure and uses a coverage penalty, encouraging generation of an output sentence that is most likely to cover all the words in the source sentence. \cite{wu2016google} The maximum length of generated headlines is 20 (not including $<SEP>$), same as the limit for headlines on Xiaohongshu. Given that all the articles within the same time step $t_x$ share the same trend text, our model can be conveniently put into usage. For social media platforms, they can update the buzzwords regularly(such as once a week) and combine user's posted headlines and buzzwords to given several candidates of personalized and trend-aware headlines for a new post. 

%% file: experiment.tex
\section{experiments}
\subsection{Experiment Settings}

\noindent\textbf{Datasets preprocessing.}
The raw dataset is described in \ref{dataset}. We preprocess the raw dataset for our experiments.
Since we focus on generating \textbf{popular} headlines, we filter those posts with less than 500 likes. 
We also filter posts whose article or headline is empty, since they lacks input or output for the headline generation task.
As described in \ref{chap:collect}, posted headlines are required to form the style text, so we filter posts whose user has no former posts, ensuring that every article has non-empty style text.
Since the styles and trends in our experiments are time-sensitive, posts in the future should be ``invisible'' to \modelname~ during training. Therefore, we use the posts in 2021 for training and those in 2022 for validation and test. 
The preprocessed dataset contains 248,311 posts in total. The train dataset consists of 150,770 posts, all of which are in posted 2021. The validation and test datasets are randomly splitted from all the posts in 2022, consist of 9,780 and 87,761 posts respectively. \\

\noindent\textbf{Baselines.}
We compare our proposed model, \modelname, with a variety of advanced baselines. They are described below:
\begin{itemize}[leftmargin=*]
    \item BART \cite{lewis2019bart} is a denoising autoencoder for pretraining sequence-to-sequence models. 
    \item PEGASUS \cite{zhang2020pegasus} is an abstractive summmarization sequence-to-sequence model pretrained with extracted gap-sentences.
    \item BRIO \cite{liu2022brio} proposed a training object that encourages coordination of probabilities and qualities among non-reference candidates generated by abstractive summarization models.
    \item Inference Time Style Control (ITSC) \cite{cao2021inference} proposed Decoder state adjustment and Word unit prediction to adjust style of summary during inference time. 
    \item TitleStylist \cite{jin2020hooks} generates style-specific headlines by combining the summarization and reconstruction tasks into a multitasking framework. 
    % \fzx{baseline}
%Furthermore, detailed case studies illustrate various diffusion patterns of specific products, which provide additional insights that will help to facilitate a full understanding of user online purchasing behaviors.
\end{itemize}

%  In PEGASUS, important sentences are removed/masked from an input document and are generated together as one output sequence from the remaining sentences. 

%  It achieves the SOTA performance on CNNDM and XSUm summarization datasets.\\

\noindent\textbf{Evaluation metric.}
To evaluate the performance of each model, we use the metric called Recall-Oriented Understudy for
Gisting Evaluation (ROUGE), which counts the number of overlapping units between the computer-generated summary to be evaluated and the ideal summary created by humans. \cite{lin2004rouge}\\

\noindent\textbf{Implementation details.}\label{setup}
During building the headline vocabulary $V$, we set $tf_{min} = 10$ and $tf_{max} = 0.01$. That is, tokens that appear less than 10 times or tokens with more than 0.01 tf are not included in $V$. 
The maximum input sequence lengths of $f_{en}(\cdot)$, $f_{se}(\cdot)$ and $f_{te}(\cdot)$ are the same, namely 512. The random seed is set to 3407 for all our experiments. 
We initialize the weights of our model \modelname~ with those of a pretrained BART model (bart-base-chinese) \cite{shao2021cpt}. 
In detail, we initialize the style extractor $f_{se}(\cdot)$, trend extractor $f_{te}(\cdot)$ and encoder $f_{en}(\cdot)$ from the pretrained encoder of the BART, but the weights are not tied during training. We initialize the decoder $f_{de}(\cdot)$ from the pretrained decoder of the same BART above. The total size of \modelname~ is about 771M.

For pertraining and training, the batch size is 64 and the updating step is 4 batches. We use the Adam optimizer \cite{kingma2014adam} with learning rate scheduling:

\begin{equation}
\label{eq:pretrain lr} 
\begin{aligned}
    lr = 2 \times 10^{-3} min(step^{-0.5}, step \cdot warmup^{-1.5})
\end{aligned}
\end{equation}
where $lr$ is the learning rate, $step$ is the number of updating steps, $warmup$ denotes the warmup steps, which is set to 100. 

To decide the appropriate number of epochs for pretraining, we pretrain \modelname~ for 5 epochs and evaluate its performance on the validation dataset at the end of each epoch. 
We then choose the pretrained model which achieves highest accuracy on the pretraining task (masked tokens prediction). The chosen pretrained model is then finetuned.

For training, to fairly measure each model's performance, we finetune all the models for 5 epochs and evaluate their performance on the validation dataset every 64 updating steps in terms of ROUGE F1 scores. 
For each model, we choose its parameters with the best ROUGE-L F1 score on the validation dataset, then evaluate its performance on the test dataset, reporting ROUGE-1/2/L F1 scores.  

\subsection{Results}
We present the performance of all the models on headline generation task on the test dataset in Table \ref{tab:Models}.
Overall, \modelname~ achieves the best performance, brings average increase of 9.8\%, 12.3\% and 9.9\% in terms of ROUGE-1, ROUGE-2 and ROUGE-L compared with the baselines. 
But how could we relate the ROUGE scores with \modelname~'s ability of modeling the styles and trends? 
To understand this, let's think about the process of creating a new headline from the perspective of the user. 
The observation has demonstrated significant similarity between headlines of the same user, supporting the existence of personal styles. 
In other words, a user probably tends to create a new headline that is similar to the his/her old ones.
Besides, the posts in our experiments are all popular posts, which means their headlines have great chance of containing buzzwords. 
Therefore, informed of the old headlines(personal styles) and the buzzwords(trends), \modelname~ is able to more accurately predict the new headline. 

To generate headlines with one certain style, ITSC requires a trained style scorer and a vocabulary of the style for the Decoder state adjustment and the Word Unit Prediction respectively. 
Since there are more than 40,000 users, which means this method demands more than 40,000 trained style scorers and vocabularies, the time cost is not affordable.

Similarly, TitleStylist has Style-Dependent Query Transformation
and Layer Normalization in its model, which means more than 40,000 versions of them require to be trained. 
Therefore, the time cost of TitleStylist is also not affordable. 
% \fzx{baseline not affordable}

The performance of PEGASUS is worse than BART. 
We think this may result from the pretraining task of PEGASUS-removing/masking important sentences from the input document and generating them as one output sequence from the remaining sentences, similar to an extractive summary. 
PEGASUS thus tends to generate the headline by concating important tokens from the input document (article). 
However, unlike news, the article and its headline on social medias are not so tightly associated, which means the way in which PEGASUS generate headlines may not apply under this circumstance. 
The performance of BRIO is better than BART, since it is able to assign higher estimated probability to the better candidate summary during inference. 
In conclusion, due to the ability of modeling styles and trends, \modelname~ achieves the best performance compared to other baselines.

\begin{table}[t!]
 \setlength{\tabcolsep}{15pt}
 \caption{ \label{tab:Models}
Experimental results on the test dataset of headline generation task.
R-1/2/L are the ROUGE-1/2/L F1 scores. TLE! means the time limit of 1000 hours exceeds. 
% \fzx{time out}
}
\centering 
\begin{tabular}{lcccr}
\toprule
\textbf{Model} & \textbf{R-1} & \textbf{R-2} & \textbf{R-L} \\
\midrule
 BART           & 28.33 & 15.19 & 25.09  \\
 PEGASUS        & 27.31 & 14.23 & 24.21  \\
 BRIO           & 28.79 & 15.32 & 25.32  \\
 \midrule 
 ITSC           &TLE!    &TLE!     &TLE!     \\
 TitleStylist   &TLE!    &TLE!     &TLE!       \\
\midrule
 \textbf{\modelname}     & \textbf{30.88} & \textbf{16.73} & \textbf{27.33} \\
\bottomrule
\end{tabular}
\end{table}

\subsection{In-depth Analysis of \modelname}

\textbf{The effect of different components.}
\modelname~ consists of two preference-Extractors, Transformer based encoder and decoder. 
To investigate whether the two preference-Extractors, namely style extractor and trend extractor, actually work and how they influence the performance of \modelname, we conduct ablation studies by removing one of them at a time, and evaluate the performances of these variants of \modelname~ in Table \ref{tab:ablation}

For every variant, we use the same setup of pretraining and training as \modelname. 
Style extractor $f_{se}(\cdot)$ and trend extractor $f_{te}(\cdot)$ are designed to capture the styles and trends respectively. 
We compare the performance of \modelname~ without one of these components. 
Removing style extractor leads to a decline of 10.9\%, 13.3\% and 11.2\% in terms of ROUGE-1, ROUGE-2 and ROUGE-l.
Removing trend extractor leads to a decline of 2.2\%, 4.7\% and 2.4\% in terms of ROUGE-1, ROUGE-2 and ROUGE-l.
We notice that the variant with only $f_{te}(\cdot)$ perform worse than BART. However, using $f_{se}(\cdot)$ and $f_{te}(\cdot)$ together does improve the proposed model compared to using $f_{se}(\cdot)$ only. We think only using $f_{te}(\cdot)$ may overemphasize the importance of buzzwords, thus misleading the proposed model. But when combined with $f_{se}(\cdot)$, the misleading is corrected under the supervision of personal styles. In other words, the proposed model pays more attention to the trends that truly cooperate with personal styles.

To summarize, according to the above discussions: (1) combining style extractor and trend extractor improves the proposed model to a large extent (2) only using style extractor can also improve the proposed model to a smaller extent, but using only trend extractor has the negative effect.

\begin{table}[t!]
 \caption{ \label{tab:ablation} 
Ablation study on \modelname. ``w/o'' means \modelname~ without a certain component.
R-1/2/L are the ROUGE-1/2/L F1 scores.
}
\centering 
\begin{tabular}{lccc}
\toprule
\textbf{Model} & \textbf{R-1} & \textbf{R-2} & \textbf{R-L} \\
\midrule
 w/o Style Extractor  & 27.52 & 14.5 & 24.28  \\
 w/o Trend Extractor  & 30.21 & 15.94 & 26.68  \\
\midrule
 \textbf{\modelname}     & \textbf{30.88} & \textbf{16.73} & \textbf{27.33} \\
\bottomrule
\end{tabular}
\end{table}

\begin{table}[t!]
\caption{ \label{tab:pretrain} 
Performances of unpretrained and pretrained \modelname. R-1/2/L are the ROUGE-1/2/L F1 scores.
}
\centering 
\begin{tabular}{lcccr}
\toprule
\textbf{Pretrained} & \textbf{R-1} & \textbf{R-2} & \textbf{R-L} \\
\midrule
 No  & 29.57 & 15.39 & 26.04  \\
\midrule
 Yes  & \textbf{30.88} & \textbf{16.73} & \textbf{27.33} \\
\bottomrule
\end{tabular}
\end{table}

\begin{table*}
\small
\caption{ \label{tab:style case}
Example headlines with strong personal styles. REF denotes the headlines created by the user. USER A prefers the pattern ``motd |'' in his/her headline, while USER B prefers ``ootd'' .
CH means the original Chinese headline, while EN means the translated headline used for display. 
% \fzx{translated}
}
\centering 
\begin{tabular}{ c | c | p{5cm} | p{5cm} | p{5cm} }
\toprule
  USER & Case  & REF & MEBART & BART \\ 
\hline  
A       & a1(CH)  
        & \begin{CJK}{UTF8}{gbsn}\textcolor{red}{motd｜}今天是chanel girl \end{CJK} 
        & \begin{CJK}{UTF8}{gbsn}\textcolor{red}{motd｜}秋冬氛围感妆容 \end{CJK} 
        & \begin{CJK}{UTF8}{gbsn}今天是小香味的一天 \end{CJK}  \\

        & a1(EN)  
        & \textcolor{red}{motd |} Today is the chanel girl
        & \textcolor{red}{motd |} Autumn/Winter Ambiance Makeup
        & Today is a day of small fragrance \\
\hline       
A       &a2(CH)  
        & \begin{CJK}{UTF8}{gbsn}\textcolor{red}{motd｜}冬日清冷易碎感 \end{CJK} 
        & \begin{CJK}{UTF8}{gbsn}\textcolor{red}{motd｜}清冷易碎感妆容 \end{CJK} 
        & \begin{CJK}{UTF8}{gbsn}清冷易碎感碎钻妆容\end{CJK}  \\

        &a2(EN)  
        & \textcolor{red}{motd |} Winter cool and fragile sense
        & \textcolor{red}{motd |} Cool and fragile sense makeup
        & Cool and fragile sense of broken diamond makeup \\
\hline  
A       & a3(CH)  
        & \begin{CJK}{UTF8}{gbsn}\textcolor{red}{motd｜}浅画一个春天 \end{CJK} 
        & \begin{CJK}{UTF8}{gbsn}\textcolor{red}{motd｜}绿野仙踪 \end{CJK} 
        & \begin{CJK}{UTF8}{gbsn}绿野仙踪 | 春天来了 \end{CJK}  \\

        &a3(EN)  
        & \textcolor{red}{motd |} Simply painting a spring
        & \textcolor{red}{motd |} Green Forest Style
        & Green Forest Style | Spring is coming \\
\hline  
B       &b1(CH)  
        & \begin{CJK}{UTF8}{gbsn}\textcolor{red}{ootd｜}早春甜蜜撞色混搭 \end{CJK} 
        & \begin{CJK}{UTF8}{gbsn}\textcolor{red}{ootd｜}春夏撞色穿搭 \end{CJK} 
        & \begin{CJK}{UTF8}{gbsn}早春穿搭｜今年春夏季节绿色系和粉色系都 \end{CJK}  \\

        &b1(EN)  
        & \textcolor{red}{ootd |} Early spring sweet  contrast color dressing match
        & \textcolor{red}{ootd |} Spring/Summer contrast color dressing
        & Early spring dressing | This spring and summer season green and pink are \\
\hline       
B       & b2(CH)  
        & \begin{CJK}{UTF8}{gbsn} \textcolor{red}{ootd｜} 早春草莓泡泡糖色系好嫩哟 \end{CJK} 
        & \begin{CJK}{UTF8}{gbsn} \textcolor{red}{ootd｜} 粉粉嫩嫩的休闲风穿搭 \end{CJK} 
        & \begin{CJK}{UTF8}{gbsn}早春穿搭｜芭比粉真的是千禧年y2k辣妹艺 \end{CJK} \\
        
        & b2(EN)
        & \textcolor{red}{ootd |} Early spring strawberry bubblegum color system so tender yo
        & \textcolor{red}{ootd |} Pink casual style dressing
        & Early spring wear | Barbie pink is really a millennial y2k hot girl art \\
\hline  
B       &b3(CH)  
        & \begin{CJK}{UTF8}{gbsn} 《我喜欢把衣服穿出花样》 \end{CJK} 
        & \begin{CJK}{UTF8}{gbsn} \textcolor{red}{ootd｜} 一周穿搭不重样 \end{CJK} 
        & \begin{CJK}{UTF8}{gbsn}早春穿搭｜美式复古风 \end{CJK}  \\

        &b3(EN)  
        & I like to wear clothes out of the box
        & \textcolor{red}{ootd |} Weekly outfits without repeating
        & Early spring wear | American vintage style \\
        
\bottomrule
\end{tabular}
\end{table*}

\begin{table*}
\small
\caption{ \label{tab:trend case}
Example headlines with time-related words. REF denotes the headlines created by the user. 
The English headlines are translated from the original Chinese headlines for display.
% \fzx{translated}
}
\centering 
\begin{tabular}{ c|p{5cm}|p{5cm}| p{5cm}  }
\toprule
    TIME & REF & MEBART & BART \\ 
\hline  
 03/2022 \textcolor{red}{(spring)}
        & \parbox{4cm}{\begin{CJK}{UTF8}{gbsn}\textcolor{red}{春夏}小裙子购物分享我可真是爱惨啦！！！\end{CJK} }
        & \parbox{4cm}{\begin{CJK}{UTF8}{gbsn}\textcolor{red}{春夏}小裙子合集来啦~\end{CJK} }
        & \parbox{4cm}{\begin{CJK}{UTF8}{gbsn}小裙子合集来啦！快来抄作业！\textcolor{red}{(none)}\end{CJK}}  \\
        
        & \parbox{4cm}{\textcolor{red}{Spring/Summer} little dresses shopping share I really love it!!!}
        & \parbox{4cm}{ \textcolor{red}{Spring/Summer} skirt collection is coming }
        & \parbox{4cm}{The dress collection is here! Come and copy the list!\textcolor{red}{(none)}}\\
\hline  
01/2022 \textcolor{red}{(spring) }
        & \parbox{4cm}{ \begin{CJK}{UTF8}{gbsn}165/44｜全套链接！甜辣学院开衫\&毛衣\&袜\end{CJK} \textcolor{red}{(none)}}
        & \parbox{4cm}{\begin{CJK}{UTF8}{gbsn}ootd｜\textcolor{red}{早春}毛衣开衫外套穿搭合集\end{CJK} }
        & \parbox{4cm}{\begin{CJK}{UTF8}{gbsn}ootd｜毛衣开衫外套穿搭韩系学院风辣妹\textcolor{red}{(none)}\end{CJK} }\\
        
        & \parbox{4cm}{165/44 | links of the set ! Sweet \& Hot School Style Cardigan \& Sweater \& Socks \textcolor{red}{(none)}}
        & \parbox{4cm}{ootd | \textcolor{red}{Early spring} sweater cardigan jacket collection }
        & \parbox{4cm}{ootd | Sweater cardigan jacket Korean school style hot girl \textcolor{red}{(none)}}\\
\bottomrule
\end{tabular}
\end{table*}

\noindent\textbf{The effect of pretraining.}
Preference-extractors in \modelname~ are pretrained with a Masked LM task. To investigate whether pretraining actually improves the model, we finetune different \modelname~ with and without pretraining and compare their performances in Table \ref{tab:pretrain}.

We use the same setup of training as stated in \ref{setup}. 
The results show that pretrained \modelname~ outperforms the unpretraiend \modelname, bringing increase of 5.2\%, 9.4\% and 5.7\% in terms of ROUGE-1, ROUGE-2 and ROUGE-L. 
This demonstrates considerable positive effect of pretraining for the preference-extractors. 
With strengthened preference-extractors, \modelname~ is able to generate headlines with higher quality. 
To summarize, the above discussions demonstrate our pretraining task benefits the performance of \modelname.
% \fzx{only talk about good result}

\subsection{Case Study}

To get a more direct look of \modelname's ability of modeling styles, we compare several headlines for the same article respectively created by the user, \modelname~ and  BART, presented in Table \ref{tab:style case}. $(a_1,a_2,a_3)$ and $(b_1,b_2,b_3)$ are listed in the order of time. Cases in $(a_1,a_2,a_3)$ show that \modelname~ is able to imitate user's style, such as the same pattern ``motd |''. In $b_3$, \modelname~ is able to generate the pattern ``ootd |'', which can be recognized as the personal style of USER B, even when the user himself/herself actually doesn't use ``ootd |'' for the article.  

Cases in Table \ref{tab:trend case} show \modelname's ability of modeling trends. In the first case, \modelname~ is able to generate the same underlined token which means ``spring and summer'' as in the human-created headline. In the second case, informed of trend text, \modelname~ is able to generate the underlined token which means ``early spring'' even when it is not in the human-created headline. 

To summarize, these examples demonstrate that \modelname~ has the ability to model personal styles and trends.

%% file: related.tex
\section{related work}
Headline generation is a popular field of research, which can be regarded as a branch of summarization. 
Summarization works can be roughly divided into two categories - extractive methods and abstractive methods. The former mainly focuses on extracting and compressing sentences from the source document or article \cite{wong2008extractive,cohn2008sentence,nallapati2017summarunner,narayan2018ranking,liu2019fine},
while the latter is based on document comprehension, capturing the salient ideas of the source text. 
Recently, Transformer \cite{vaswani2017attention} based methods for abstractive generation of summary or headline have shown great potential
\cite{lewis2019bart,zhang2020pegasus,hasan2021xl,hasan2021crosssum,liu2022brio,xu2022sequence} . 
They are able to generate short and concise summaries or headlines that are similar to the original ones. 
And the applications of abstractive headline generation varies in different fields, including news \cite{ao2021pens,zheng2021tweet}, online ads \cite{kanungo2022cobart}, paper writing \cite{shenassa2022elmnet, xu2022sequence} and content community \cite{zhang2022diverse}. 
However, it takes more than being short and concise to be a good (popular) headline on social media. 
According to our analysis and observation, appropriately keeping personal style and following trends contribute to a popular headline. 
Although there are also works that focusing on generating personalized or stylistic headlines \cite{diaz2007user, yan2011summarize, moro2012personalized, zhang2018question, jin2020hooks, cao2021inference} , they either are limited to several specific styles or demand a large-scale corpus for the target style.
Therefore, considering the numerous users (everyone has his/her own style) and the relatively small corpus for each user (his/her posted posts), it is not appropriate to directly apply these methods to generate headlines for posts on social media. 
% \fzx{rewrite}

%% file: conclusion-and-future-work.tex
\section{Conclusion}
This paper present a novel study: automating generating popular headlines on social medias; and investigate the problem from the perspectives of trends and personal styles. 
To figure out how trends and personal styles affect the generation process and popularity of headlines, this paper collected more than 1 million posts from 42,447 thousand from public data of Xiaohongshu and conduct careful observations on the headlines of these posts. Results suggest that trends and personal styles are indeed widespread in the headlines of posts and have significant impacts on posts' popularity. 
Therefore, we present \modelname. It is a BART-based method, with two special preference-Extractors to capture trends and personal styles. 
We conduct extensive experiments on Xiaohongshu datasets. 
\modelname~achieves the best performance on the test dataset, bringing average increase of 9.9\% in terms of ROUGE-L compared with the baselines. 
Meanwhile, ablation studies demonstrate that modeling trends and personal styles indeed prompt the model's performance. 
Besides, the results also demonstrate the proposed pretraining strategy is powerful. To further study the advancement of \modelname, we present several cases. 
In summary, these results demonstrate that \modelname~indeed is an effective, trends-aware, personalized headline generation method on social medias.

%% file: appendix.tex
\newpage
\section{appendix}

\subsection{Analysis with other metrics of text similarity}\label{metric}

\begin{figure*}[!hb]
\centering
\setlength{\abovecaptionskip}{0cm} %调整图片标题与图距离
\setlength{\belowcaptionskip}{0cm} %调整图片标题与下文距离
\vspace{0cm}
\includegraphics[width=0.7\linewidth]{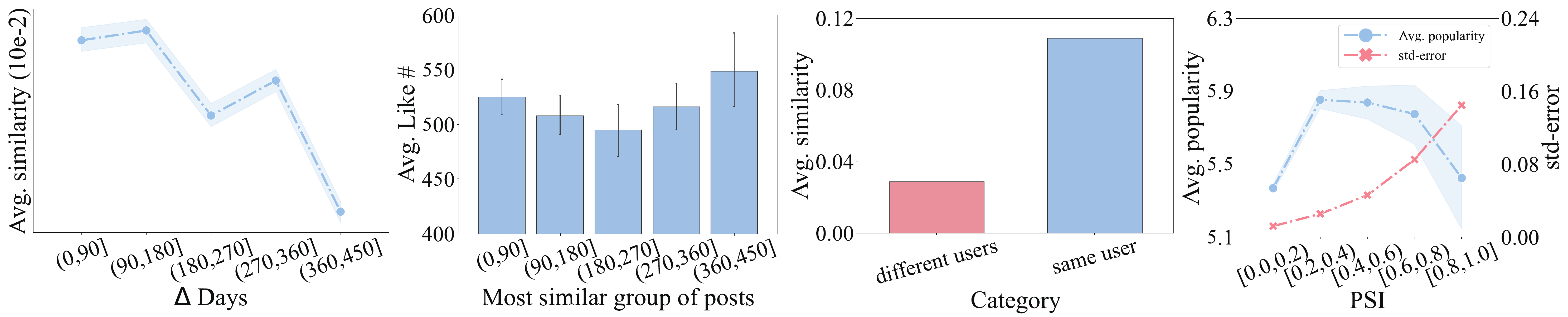}
\caption{Observations of the trends and personal styles of headlines using LCS as the metric of text similarity}
\label{fig:lcs}
\end{figure*}

\begin{figure*}[!hb]
\centering
\setlength{\abovecaptionskip}{0cm} %调整图片标题与图距离
\setlength{\belowcaptionskip}{0cm} %调整图片标题与下文距离
\vspace{0cm}
\includegraphics[width=0.7\linewidth]{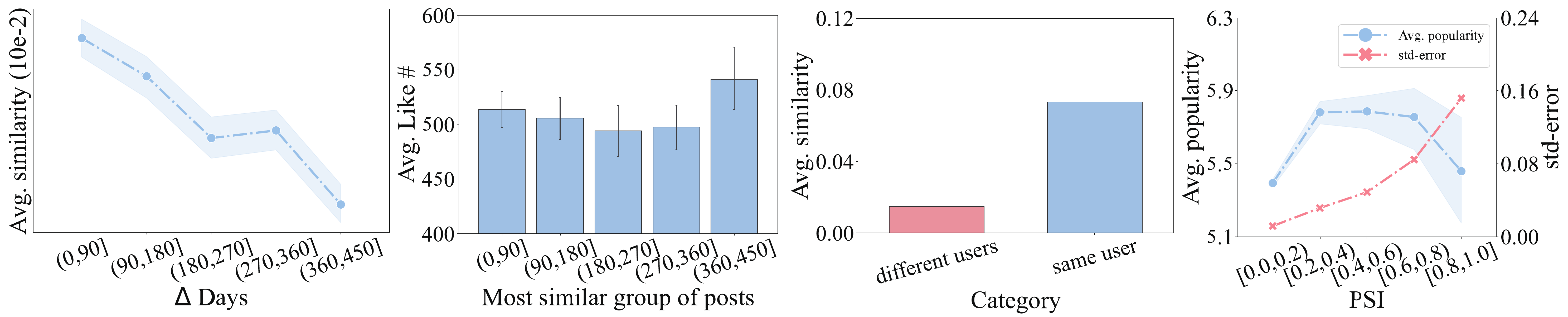}
\caption{Observations of the trends and personal styles of headlines using N-gram as the metric of text similarity.}
\label{fig:ngram}
\end{figure*}

\begin{figure*}[!hb]
\centering
\setlength{\abovecaptionskip}{0cm} %调整图片标题与图距离
\setlength{\belowcaptionskip}{0cm} %调整图片标题与下文距离
\vspace{0cm}
\includegraphics[width=0.7\linewidth]{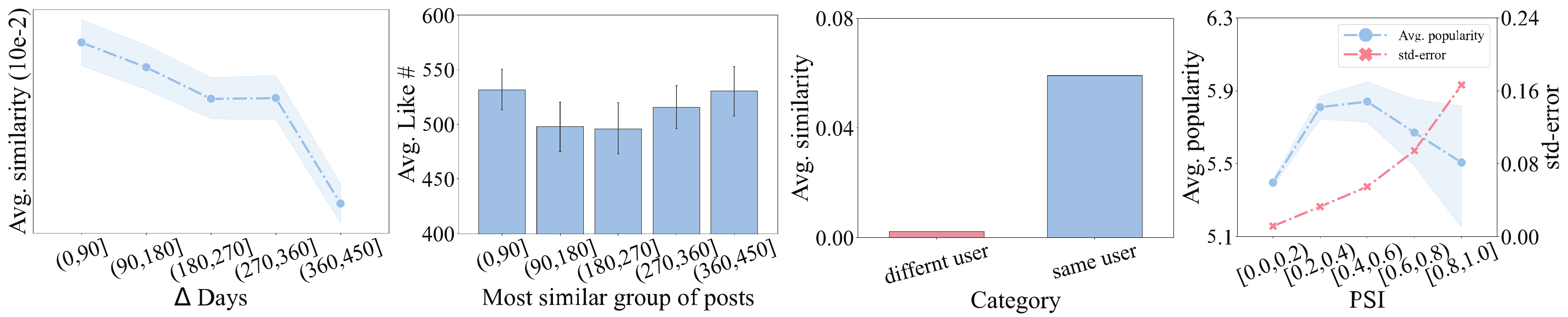}
\caption{Observations of the trends and personal styles of headlines using Cosine similarity as the metric of text similarity.}
\label{fig:cosine}
\end{figure*}

We replace \textit{Jaccard similarity} with \textit{Longest Common SubString (LCS)}, \textit{N-gram} and \textit{Cosine similarity} as the metric of text similarity.
We still sampled 50,000 headlines from the last 10 days of 2021 as our study subject and repeat our analysis of trends and personal style. 
The results in figure \figref{fig:lcs} , figure \figref{fig:ngram} and figure \figref{fig:cosine} is quite similar to \figref{fig:analysis}.
\begin{figure*}[!hb]
\setlength{\abovecaptionskip}{0cm} %调整图片标题与图距离
\setlength{\belowcaptionskip}{0cm} %调整图片标题与下文距离
\includegraphics[width=0.2\textwidth]{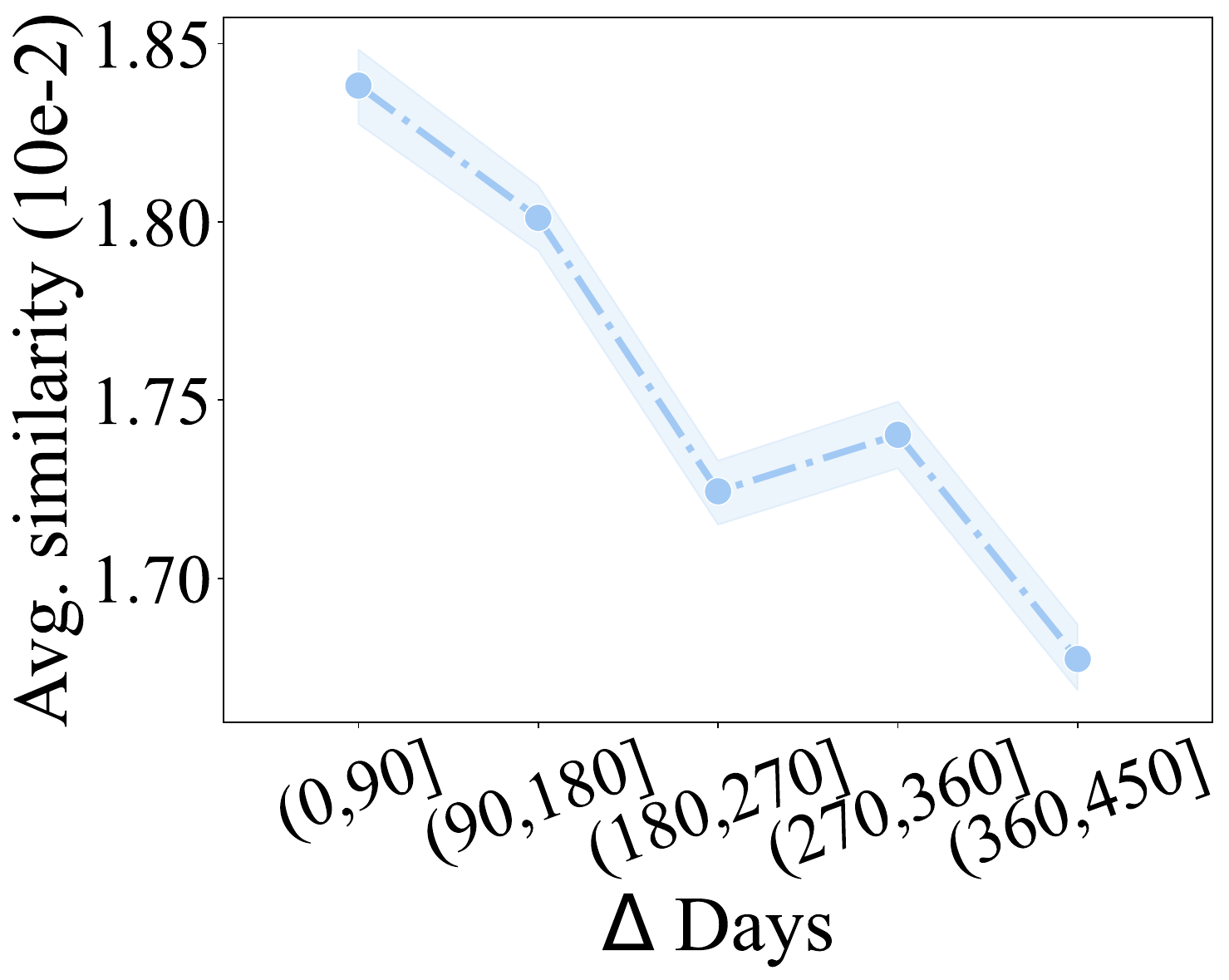}
\caption{The average similarities between the headlines from different seasons
\normalsize
}
%and new tasks arriving. When a new task arrives, this method creates a new GNN, which learns the task under the current graph structure..}
\label{fig:time1}
\end{figure*}

\begin{figure*}[!hb]
\setlength{\abovecaptionskip}{0cm} %调整图片标题与图距离
\setlength{\belowcaptionskip}{0cm} %调整图片标题与下文距离
\includegraphics[width=0.2\textwidth]{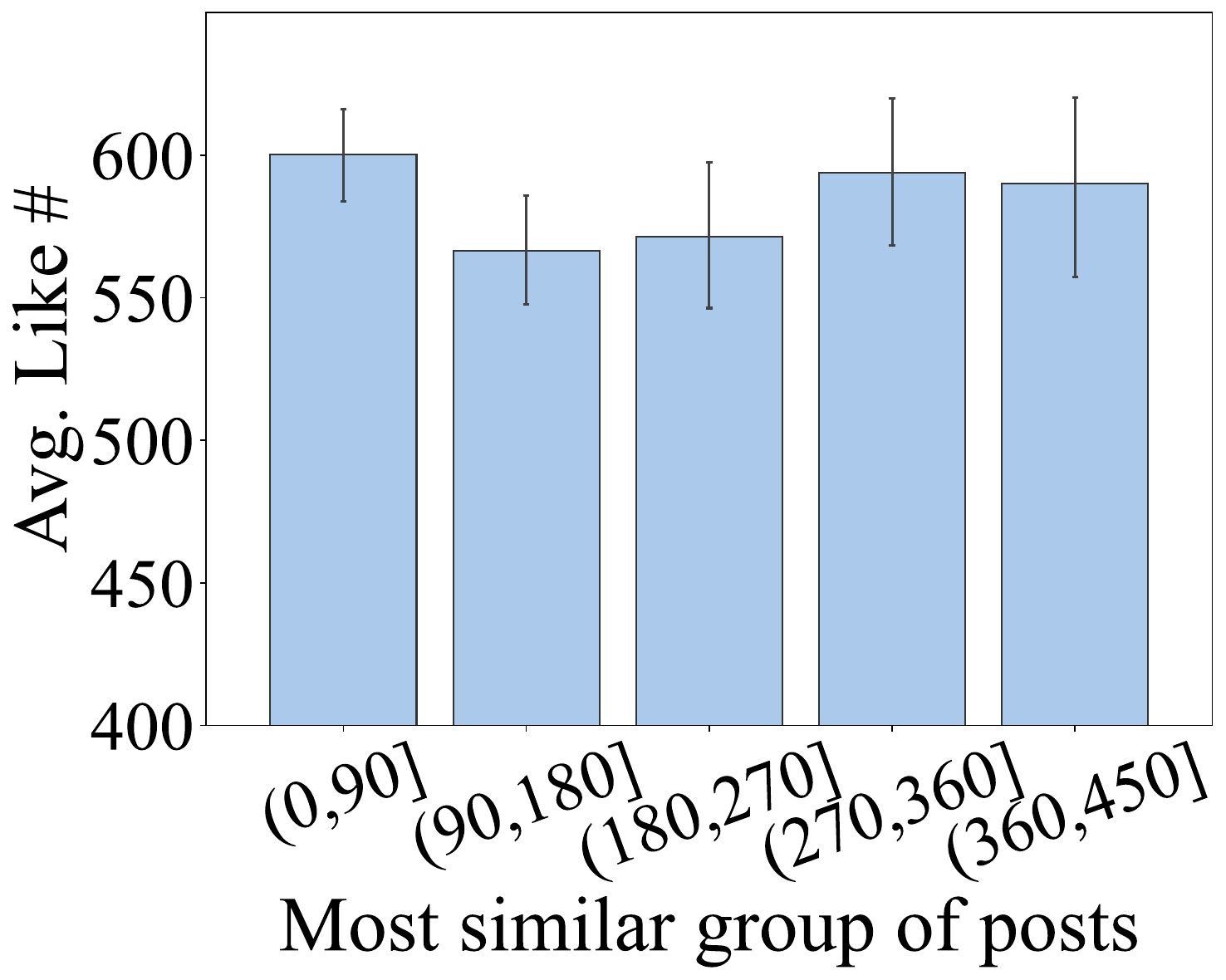}
\caption{The average likes of headlines from different season groups
\normalsize
}
%and new tasks arriving. When a new task arrives, this method creates a new GNN, which learns the task under the current graph structure..}
\label{fig:time2}
\end{figure*}
\subsection{Analysis with other time period}\label{time}

We replace our study subject with 50,000 sampled headlines from the last 10 days in the third season of 2021 and use Jaccard similarity as the metric of text similarity. 
We then repeat our analysis for trends.
The results in figure \figref{fig:time1} and figure \figref{fig:time2} is quite similar to \figref{fig:analysis}~a and \figref{fig:analysis}~b.